\documentclass[10pt,twocolumn,letterpaper]{article}

\usepackage[pagenumbers]{wacv} 

\definecolor{wacvblue}{rgb}{0.21,0.49,0.74}
\usepackage[pagebackref,breaklinks,colorlinks,allcolors=wacvblue]{hyperref}
\usepackage{amsmath, amssymb}
\usepackage[ruled,vlined,linesnumbered]{algorithm2e}
\usepackage{booktabs}    
\usepackage{multirow}
\usepackage{colortbl}    
\usepackage{xcolor}
\usepackage{graphicx}
\usepackage{subcaption}
\usepackage[capitalize]{cleveref}
\usepackage{xcolor}
\usepackage{cuted}
\usepackage{capt-of}
\usepackage{pifont}
\newcommand{\cmark}{\ding{51}} 
\newcommand{\xmark}{\ding{55}} 

\def\wacvPaperID{334} 
\def\confName{WACV}
\def\confYear{2027}

\title{Preserve the Hard, Regenerate the Rest: Uncertainty-Guided Synthetic Training Data Augmentation with Diffusion Models} 

\author{
    Nikolai Röhrich\textsuperscript{1,2} \hspace{0mm}
    Julian Gleißner\textsuperscript{1} \hspace{0mm}
    Ahmed H. A. Ibrahim\textsuperscript{1,3} \hspace{0mm}
    Silvan Mertes\textsuperscript{4} \hspace{0mm}
    Tobias Huber\textsuperscript{1,3}\thanks{Corresponding author: {\tt\small tobias.huber@thi.de}}
    \vspace{2mm} \\
    {\small 
    \textsuperscript{1}XITASO GmbH \hspace{0mm}
    \textsuperscript{2}Zuse School of Excellence in Reliable AI \hspace{0mm}
    \textsuperscript{3}Technische Hochschule Ingolstadt \hspace{0mm}
    \textsuperscript{4}Technische Hochschule Augsburg
    }
}

\begin{document}
\maketitle

\begin{strip}
    \vspace{-10mm}

    \centering
    \includegraphics[width=0.98\textwidth]{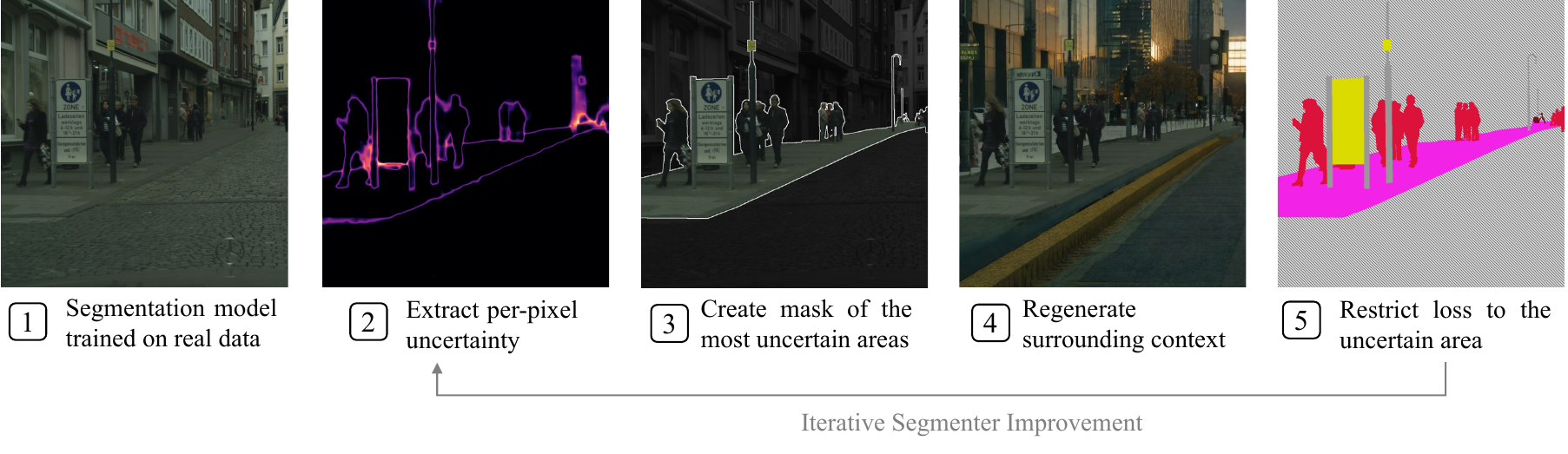}
    \captionof{figure}{\textbf{Uncertainty-Guided Context Synthesis.} We augment real samples by computing the current segmenter's per-pixel predictive entropy and inpaint a fresh visual context around hard-to-classify regions. Our method transcends prior region selection heuristics by allocating synthetic budget on informative regions rather than always choosing background or foreground regions.}
    \label{fig:method_figure}
\end{strip}

\begin{abstract}
Semantic segmentation models struggle with data sparsity and rare or visually diverse regions, e.g., dense regions or small objects in aerial or autonomous mobility data. While synthetic augmentation is an appealing solution, directly generating new labeled data risks misalignment of labels and generated pixels. Existing solutions to this problem often rely on external models, or employ coarse heuristics such as indiscriminately augmenting all foreground objects or entire backgrounds, which wastes capacity on uninformative pixels. To address this, we propose an uncertainty-guided synthetic context augmentation strategy that strictly preserves label validity and efficiently maximizes pixel informativeness per synthetic sample – no external guardrails required. Using a baseline segmenter's predictive entropy, we identify uncertain semantic regions and inpaint only the complementary visual context.
When fine-tuning the segmenter on this synthetic data, we compute the loss only over the original pixels, excluding inpainted regions. This focuses learning on the unmodified, uncertain regions while presenting them in novel contexts. We demonstrate substantial mIoU gains on Cityscapes, UAVID, and BDD100K with the largest gains on rare and difficult classes such as buses, trains, or (from the aerial perspective) cars. Our results demonstrate that uncertainty-guided context augmentation is a highly effective lever to improve segmentation performance on complex datasets, with \href{https://github.com/XITASO/Preserve-the-Hard-Regenerate-the-Rest}{code provided}.
\end{abstract}
    
\section{Introduction}
\label{sec:intro}

Data sparsity remains a core challenge for semantic segmentation models, particularly for datasets with rare classes and visually diverse regions \cite{feng2020deep, janai2020computer}. Examples include dense scenes and small objects in datasets with a high level of local detail. Such data are particularly tedious to annotate: For instance, annotating a single urban driving scene takes up to 90 minutes \cite{cordts2016cityscapes}. Real-world data are also highly imbalanced, and consequently, the most critical classes are often those with the fewest labeled pixels \cite{cui2019class,lin2017focal}. 

The rise of generative diffusion models \cite{ho2020denoising,rombach2022high, peebles2023scalable, podell2024sdxl} has made synthetic data an attractive lever to close this gap. Two principal strategies have emerged here: 
The first generates images and labels jointly \cite{wu2023diffumask, wu2023datasetdm,yang2023freemask, nguyen2023dataset}. While conceptually appealing, these approaches are constrained by a label–pixel mismatch problem: the diffusion model must either invent labels for the pixels it generates, or generate pixels that faithfully match existing labels. In neither case, there is a guarantee for exact label-to-pixel correspondence.

\begin{figure}[t]
    \centering
    \includegraphics[width=\linewidth]{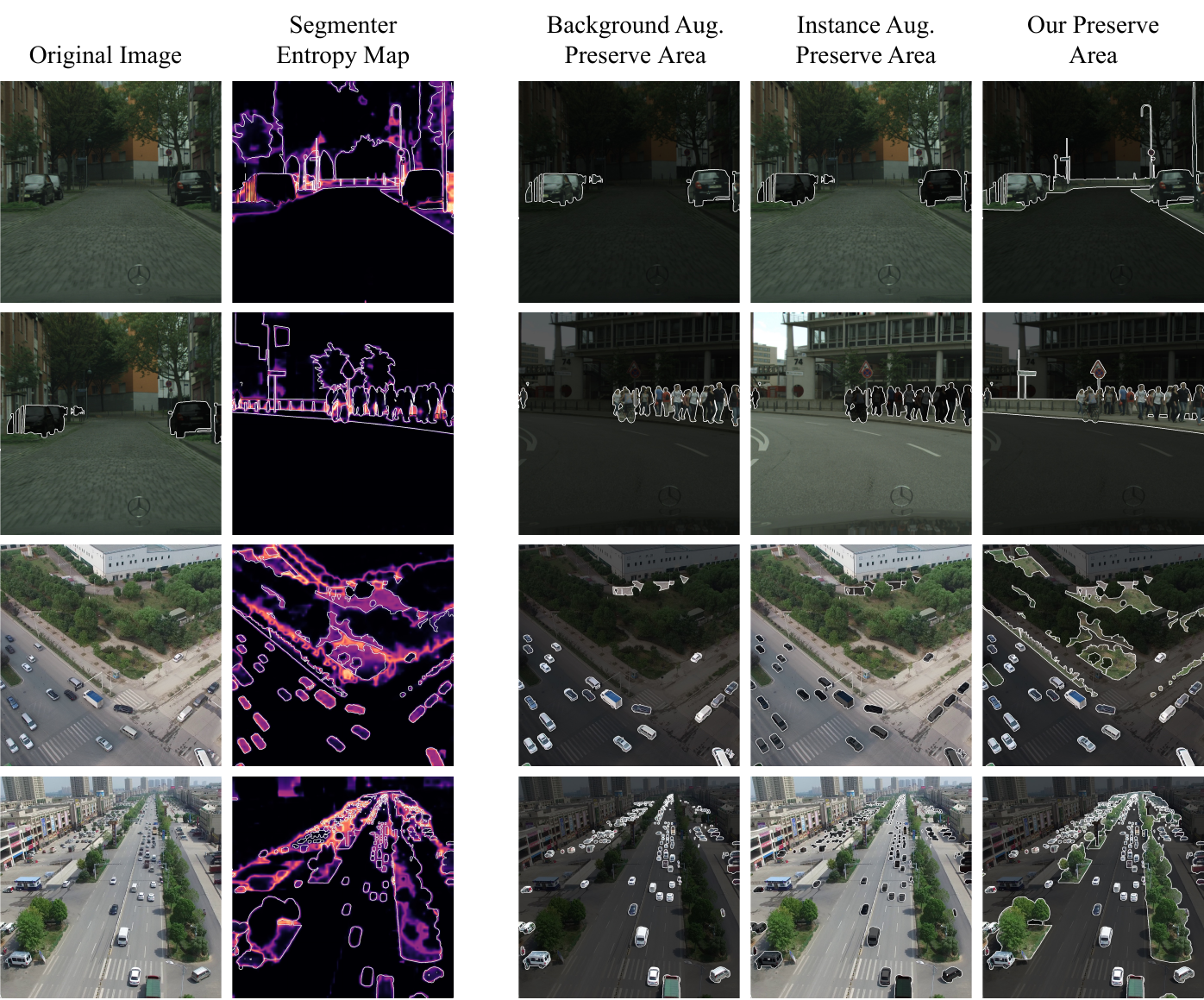}
    \caption{\textbf{Uncertainty-based region selection transcends the foreground-background contradiction.} We argue that the augmented regions should be based on uncertainty, rather than always selecting either the entire foreground \cite{kupyn2024dataset} or background \cite{li2024simple}.}
    \label{fig:foreground_background_contradiction}
\end{figure}

A second, more conservative line of work sidesteps this problem by partially regenerating real images while keeping the original annotations intact \cite{zhao2023x,islam2024diffusemix, kupyn2024dataset,li2024simple}. Here, the literature has converged on coarse spatial heuristics: some methods regenerate the foreground objects of the scene while preserving the background \cite{kupyn2024dataset}, while others do the exact opposite, regenerating the entire background while preserving the foreground objects \cite{li2024simple}. Strikingly, these two heuristics reach opposite empirical conclusions. \citet{kupyn2024dataset} attribute their performance gains to the visual diversity injected by redrawing foreground objects, whereas \citet{li2024simple} report that background augmentation outperforms object augmentation by a wide margin. We argue that this contradiction is not a matter of who is right: it demonstrates that the foreground/background axis is the wrong axis along which to allocate the synthetic-data budget. The right axis, we argue, is informational, not spatial: The pixels that deserve to be preserved are those on which the current segmenter is most uncertain (see \Cref{fig:foreground_background_contradiction}). Conversely, regions suitable for regeneration are those which the segmenter is already confident in and which therefore provide limited training signal, regardless of whether they belong to certain spatial categories \cite{shrivastava2016training,yoo2019learning}. 

Building on this insight, we propose an uncertainty-targeted synthetic context augmentation strategy. Given a segmenter trained on the available real data, we compute its per-pixel predictive entropy \cite{gal2016dropout,mackowiak2018cereals} and aggregate it over each ground-truth semantic region. The regions with the highest mean entropy are selected greedily to form a preserve mask of uncertain pixels. Diffusion-based inpainting \cite{lugmayr2022repaint,podell2024sdxl} is then applied to the complement of this mask, generating a novel visual context around the preserved region. Crucially, the original pixels inside the mask remain bitwise identical, so the labels they carry are guaranteed valid. 

This design has three consequences: First, label noise is eliminated by construction rather than mitigated by external models. We require no ControlNet \cite{zhang2023adding}, no edge or depth conditioning, no mask-refinement network, and no auxiliary VQA prompts \cite{li2022blip} to keep the inpainter on-label. Second, by computing the loss only for the preserved uncertain pixels, the training is driven by the most informative regions, while inpainted pixels serve as novel context.
Finally, this design is naturally iterative, since the preserve mask is defined by the \emph{current} segmenter's predictive entropy.
Improving the segmenter shifts uncertainty onto new regions and additional rounds of selection-and-inpainting target exactly those regions. Repeating the loop turns one-shot augmentation into an active-learning-inspired cycle \cite{settles2009active, gal2017deep,yoo2019learning} in which the diffusion inpainter (rather than a human annotator) supplies the answer to each round's query, with gains accumulating until uncertainty saturates.

We evaluate our method on semantic segmentation for three benchmarks with a high level of complexity and local diversity: Cityscapes \cite{cordts2016cityscapes} and BDD100K \cite{yu2020bdd100k} (urban driving), and UAVID \cite{lyu2020uavid} (aerial urban scenes). On all benchmarks, mean IoU improves substantially, with gains concentrated on rare and safety-critical classes. Contributions:
\begin{itemize}
    \item \textbf{Strictly label-preserving augmentation.} Synthetic pixels are explicit ignore regions, so our synthetic data injects context diversity without injecting any label noise, and without requiring any external guardrails.
    \item \textbf{Uncertainty-targeted region selection.} We allocate synthetic-data capacity along the informational axis rather than the spatial one, overcoming the foreground-vs-background contradiction in prior work (see \Cref{fig:foreground_background_contradiction}).
    \item \textbf{Empirical gains.} We achieve $+3.68 / +2.64 / +1.57$ mIoU on $\text{UAVID } / \text{ Cityscapes } / \text{ BDD}$ compared to the real-data only segmenter, and more than double the gain of the best augmentation baseline on Cityscapes, supporting that the right unit of augmentation is uncertainty.
    \item  \textbf{An iterative active-learning loop.} Because selection is driven by the current model's uncertainty rather than a fixed spatial rule, multiple iterations can re-target the segmenter's latest failure modes. This compounds gains in a way one-shot methods cannot: on Cityscapes the \textit{train} class IoU, which the first iteration barely moves ($+1.1$), later climbs to $+9.2$ once uncertainty concentrates on it.
\end{itemize}

\section{Related Work}
\label{sec:related}

\paragraph{Predictive Uncertainty in Segmentation.} Active learning~\cite{settles2009active,gal2017deep,yoo2019learning} traditionally uses model uncertainty to query the most informative unlabelled samples for human annotation. Our method draws on the same underlying signal but repurposes it to decide which pixels are worth surrounding with fresh synthetic visual context. Closest to our work is \textit{Uncertainty-Aware ControlNet}~\cite{niemeijer2025uncertainty}, which uses a per-pixel entropy map as a conditioning input to a second ControlNet branch and generates full synthetic images to bridge a domain gap, and \textit{Active Learning Inspired ControlNet Guidance}~\cite{kniesel2025active}, which uses prediction entropy for classifier guidance, thus requiring gradient flow through each diffusion step while also employing a second ControlNet branch. The role of entropy in those pipelines is orthogonal to ours: they generate whole images conditioned on entropy, whereas we use entropy to choose what to preserve and leave the actual generative process unconditioned on the segmenter. At the same time, our method refrains from auxiliary models and generates using only an out-of-the-box inpainting model.
 
\paragraph{Diffusion-Based Data Synthesis.} A growing body of work uses diffusion models~\cite{rombach2022high,podell2024sdxl,peebles2023scalable} to generate synthetic data for dense-prediction tasks. The dominant strategy is \emph{generate-then-label}: the diffusion model produces an image and a paired label map is recovered either from the model's internal cross-attention representations (\textit{DiffuMask}~\cite{wu2023diffumask}, \textit{DatasetDM}~\cite{wu2023datasetdm}) or by conditioning generation directly on a target mask (\textit{FreeMask}~\cite{yang2023freemask}, \textit{DatasetDiffusion}~\cite{nguyen2023dataset}). Closely related lines synthesize data for classification using diffusion priors~\cite{trabucco2024effective}. All such methods inherit a structural label--content mismatch: because labels are either inferred from generated pixels or imposed on them, there is no guarantee that the two actually agree. 
 
A second branch of generative augmentation targets object detection rather than semantic segmentation, leveraging \textit{ControlNet}~\cite{zhang2023adding} to inject structural cues into the generation process. \cite{fang2024data} conditions Stable Diffusion on detection-friendly layouts to produce additional training images, and \textit{ReCon}~\cite{zhu2026recon} extends this idea with region-controllable rectification and alignment to better preserve bounding-box semantics. These methods operate at the granularity of bounding boxes rather than dense per-pixel labels, and rely on auxiliary structural networks to keep the generation on-task.

\paragraph{Preserve-and-Regenerate Augmentation.} \textit{InstanceAugmentation}~\cite{kupyn2024dataset} targets several tasks including semantic segmentation and redraws every annotated object in the scene while keeping the background intact. To prevent generated objects from drifting away from the original masks, the method composes a stack of auxiliary networks: ControlNets conditioned on depth and edge maps, a BLIP-VQA \cite{li2022blip} model, and a TRACER-7 segmentation network \cite{lee2022tracer}. \textit{Simple Background Augmentation~\cite{li2024simple}}, targeting instance segmentation and object detection, takes the opposite view: it regenerates the entire background of each image while keeping the foreground objects bitwise unchanged. The authors report that background regeneration outperforms foreground regeneration by a wide margin. The two papers therefore arrive at opposite empirical conclusions, a contradiction we argue arises because the foreground/background axis is not the one that matters.
 
While these works are the most relevant to our approach, it differs along four dimensions. \emph{First}, our unit of selection is not a geometric region (foreground or background) but an \emph{informationally hard} region chosen by the segmenter's own predictive entropy. \emph{Second}, where \cite{kupyn2024dataset} relies on a full stack of auxiliary models, our method achieves full label-preservation without auxiliary models. \emph{Third}, where both prior methods train on pixels they themselves regenerated (and therefore inherit label-content mismatch) we mark every generated pixel as ignore (\cref{eq:syn-label}) and train only on real labels. \emph{Fourth}, both prior methods are static augmentations applied once before training. Because our selection criterion is model-dependent, our augmentation forms a closed loop that can be re-run against the improved segmenter.

\section{Method}
\label{sec:method}

Our method consists of five main steps as shown in Fig. \ref{fig:method_figure}.
It starts (1) with a segmenter that is trained on real data. 
Then (2), it identifies the regions on which that segmenter is most uncertain, (3) preserves those pixels and their labels, and (4) inpaints a fresh visual context around them with an off-the-shelf latent diffusion inpainter. Finally (5), the segmenter is fine-tuned on the resulting augmented dataset where the loss is restricted to original pixels and labels. Due to the active-learning-inspired nature of our approach, it can  be applied iteratively for further improvements.

\subsection{Setting and Preliminaries}
\label{sec:method:prelim}

Let $\mathcal{D} = \{(x_n, y_n)\}_{n=1}^{N}$ denote a semantic segmentation dataset, where each image $x \in \mathbb{R}^{3 \times H \times W}$ is paired with a dense label map $y \in \{0, 1, \ldots, C-1, \iota\}^{H \times W}$ over $C$ semantic classes and a designated ignore index $\iota$.
For any segmenter $f_\theta : \mathbb{R}^{3 \times H \times W} \to \mathbb{R}^{C \times H \times W}$, let $p_{c,i,j}(x)$ denote the softmax probability assigned to class $c$ at location $(i,j)$ and let $\Omega_y = \{(i,j) : y_{i,j} \neq \iota\}$ be the set of valid pixels in $y$.
Then, the standard pixel-wise cross-entropy loss with ignore-index masking \citep{cordts2016cityscapes} is given by
\begin{equation}
\begin{gathered}
    \mathcal{L}(\theta;\, x, y) \;=\; -\frac{1}{|\Omega_y|} \sum_{(i,j) \in \Omega_y} \log p_{y_{i,j},\, i, j}(x).
\end{gathered}
\label{eq:loss}
\end{equation}
Let $f_{\theta_0}$ be a segmenter trained on the real dataset $\mathcal{D}$ by minimizing \cref{eq:loss}.
Our goal is to improve this real-data-only model with the proposed synthetic training data augmentation. 
Thus, we refer to it as the baseline segmenter.

\subsection{Uncertain-Region Selection}
\label{sec:method:selection}

For any image $x$, we measure uncertainty through the pixel-wise Shannon entropy of the  predictions $p_{c,i,j}(x)$ of $f_{\theta_0}(x)$:
\begin{equation}
    H_{i,j}(x) \;=\; -\sum_{c=0}^{C-1} p_{c,i,j}(x)\, \log p_{c,i,j}(x)
    \label{eq:entropy}
\end{equation}

Selecting the most uncertain regions of an image at the level of individual pixels may destroy semantic coherence, since preserving only scattered pixels of an object, such as a car, erases its class-level meaning.
Thus, we aggregate uncertainty at the level of ground-truth class regions. For an image $x$ with label map $y$, let $\mathcal{C}_x = \{c \in \{0,\ldots,C-1\} : \exists (i,j) \text{ s.t. } y_{i,j} = c\}$ be the set of valid semantic classes present in the image.
For each class $c \in \mathcal{C}_x$, let $y^c \in \{0,1\}^{H \times W}$ denote its binary ground-truth mask,
\begin{equation}
    y^c_{i,j} = 
    \begin{cases} 
    1 & \text{if } y_{i,j} = c \\
    0 & \text{otherwise}.
    \end{cases}
    \label{eq:class-mask}
\end{equation}
The mean predictive entropy over class $c$ is
\begin{equation}
    \bar{H}_c(x, y) \;=\; \frac{1}{\|y^c\|_1} \sum_{i,j} y^c_{i,j}\, H_{i,j}(x),
    \label{eq:class-entropy}
\end{equation}
where, $\|y^c\|_1 = \sum_{i,j} y^c_{i,j}$ denotes the number of pixels belonging to class $c$. We rank the classes $c\in \mathcal{C}_x$ by their mean entropy $\bar{H}_c(x, y)$ and keep adding the most uncertain classes until their combined region covers a predefined fraction $\tau$ of the full image.
Let $\sigma : \{1, \ldots, |\mathcal{C}_x|\} \to \mathcal{C}_x$ be the permutation that ranks classes by region entropy (i.e., $\bar{H}_{\sigma(1)} \geq \bar{H}_{\sigma(2)} \geq \ldots \geq \bar{H}_{\sigma(|\mathcal{C}_x|)}$).
Then we define our preserve mask as the sum of the most uncertain class masks,
\begin{equation}
    M(x,y;\tau) = \sum_{j=1}^{k} y^{\sigma(j)},
    \label{eq:preserve-mask}
\end{equation}
where $k$ is the smallest index such that $\tfrac{1}{HW}\bigl|\mathcal{M}(x,y;\tau) \bigl|_1 \;\geq\; \tau$.
$M(x,y;\tau)$ is itself binary since the class masks are disjoint.
The threshold $\tau \in (0, 1]$ controls a single, intuitive trade-off: smaller $\tau$ concentrates the synthetic-context budget on the hardest few classes per image, while larger $\tau$ preserves more of the original image at the cost of less novel context. $\tau$ is introduced as a hyperparameter and we study its effect in \cref{sec:experiments}.

By construction, $\mathcal{M}$ is image-adaptive: in an urban scene where the baseline confuses bus and train pixels, those pixels are preserved.
In a scene where the baseline is uncertain along a wall--fence boundary, those pixels are preserved instead. 
The selector therefore allocates the augmentation budget along an informational axis rather than the foreground/background axis of prior work~\cite{kupyn2024dataset,li2024simple}.

\begin{algorithm}[t]
\SetAlgoLined
\DontPrintSemicolon
\KwIn{Real dataset $\mathcal{D}$; baseline segmenter $f_{\theta_0}$; inpainter $G$; text prompt $t$; area threshold $\tau$; ignore index $\iota$.}
\KwOut{Improved segmenter $f_{\theta_1}$.}
\ForEach{$(x, y) \in \mathcal{D}$}{
    $p \gets \operatorname{pixel-wise \, softmax}\bigl(f_{\theta_0}(x)\bigr)$ 
    
    $H_{i,j} \gets -\sum_{c=0}^{C-1} p_{c,i,j} \log p_{c,i,j}$ 
    
    \ForEach{class $c \in \mathcal{C}_x$}{
        $y^c \gets \mathbf{1}[\, y = c\, ]$\;
        $\bar{H}_c \gets \tfrac{1}{\|y^c\|_1} \sum_{i,j} y^c_{i,j} H_{i,j}$ 
    }
    Sort $\mathcal{C}_x$ in descending order of $\bar{H}_c$, yielding $\sigma$.\;
    $M \gets \mathbf{0}^{H \times W}$;\; $k \gets 0$\;
    \While{$\tfrac{1}{HW}\|M\|_1 < \tau$}{
        $k \gets k + 1$;\; $M \gets M + y^{\sigma(k)}$\;
    }
    $x_{\mathrm{gen}} \gets G(x,\, \mathbf{1} - M,\, t)$ 
    
    $\tilde{x} \gets (\mathbf{1} - M) \odot x_{\mathrm{gen}} + M \odot x$ 
    
    Construct $\tilde{y}$ from $y$ and $M$.\;
    
    $\mathcal{D}_{\mathrm{syn}} \gets \mathcal{D}_{\mathrm{syn}} \cup \{(\tilde{x}, \tilde{y})\}$\;
}
$f_{\theta_1} \gets \text{fine-tune $f_{\theta_0}$ on $\mathcal{D} \cup \mathcal{D}_{\mathrm{syn}}$}$\;
\Return{$f_{\theta_1}$}\;
\caption{A single iteration of our uncertainty-informed training data generation.}
\label{alg:method}
\end{algorithm}

\subsection{Context Inpainting with Pixel Preservation}
\label{sec:method:inpaint}

Given an image $x$ and its preserve mask $M$, we generate a synthetic counterpart by inpainting only the complementary region $\bar{M} = \mathbf{1} - M$. Let
\begin{equation*}
    G : \mathbb{R}^{3 \times H \times W} \times \{0,1\}^{H \times W} \times \mathcal{T} \;\to\; \mathbb{R}^{3 \times H \times W}
\end{equation*}
be a pre-trained latent diffusion inpainter that takes an image, a binary inpainting mask, and a text prompt $t \in \mathcal{T}$, and returns an image in which the masked region is regenerated. We apply $G$ to the complement $\bar{M}$ of the preserve mask $M$,
\begin{equation}
    x_{\mathrm{gen}} \;=\; G\bigl(x,\, \bar{M},\, t\bigr),
    \label{eq:inpaint}
\end{equation}
using a single generic, dataset-level prompt $t$ (e.g.\ describing the imaging domain, see Appendix \ref{sec:supp_prompts}).
In contrast to prior preserve-and-regenerate methods~\cite{kupyn2024dataset,li2024simple}, we deliberately employ no class-specific prompting, no ControlNet \cite{zhang2023adding} conditioning, and no mask-refinement.
To guarantee label validity, we only utilize the preserve mask itself and two generic post-processing steps described below. 

\subsubsection{Paste-Back to Preserve Uncertain Pixels.} 
\begin{figure}[t]
    \centering
    \includegraphics[width=\linewidth]{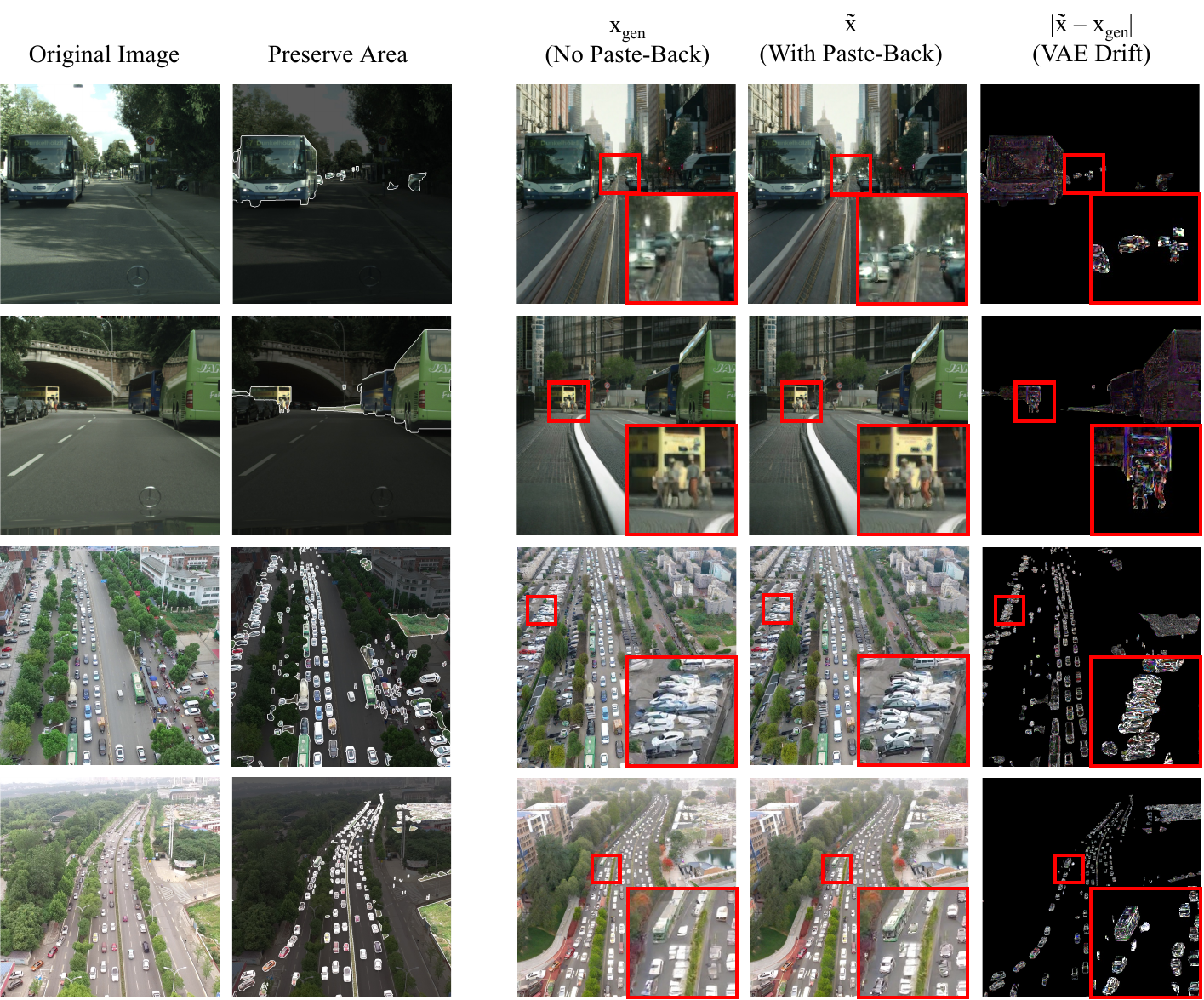}
    \caption{\textbf{Pixel-exact paste-back preserves label validity.}}
    \label{ref:teaser_figure}
\end{figure}

Latent diffusion inpainters operate by encoding the input image into a VAE latent, modifying the noised latent through the masked region, and decoding the result back to pixel space~\cite{rombach2022high}.
This encode--decode roundtrip is not lossless: even pixels outside $\bar{M}$, which the inpainter is not supposed to change, undergo small reconstruction errors \cite{lugmayr2022repaint, avrahami2023blended} that perturb their values (see \Cref{ref:teaser_figure}). 
This is especially problematic for uncertain regions which are often small and depict rare objects, leading to larger reconstruction errors. 
If left uncorrected, this drift would silently invalidate the labels we are trying to preserve. 
We therefore perform an explicit, pixel-exact paste-back step that restores the original values inside $M$:
\begin{equation}
    \tilde{x} \;=\; \bar{M} \odot x_{\mathrm{gen}} \;+\; M \odot x,
    \label{eq:paste-back}
\end{equation}
where $\odot$ denotes per-pixel multiplication broadcast over the channel dimension. I.e., the resulting image $\tilde{x}$ is bitwise identical to $x$ inside $M$ and identical to $x_{\mathrm{gen}}$ outside it.

\subsubsection{Synthetic Label Maps that Ensure Label Validity} 
To use the generated images for fine-tuning of $f_{\theta_0}$, we construct a synthetic dataset $\mathcal{D}_{\mathrm{syn}}$.
Given a real sample $(x,y)\in \mathcal{D}$, we create a synthetic image $\tilde{x}$ and define a corresponding label map $\tilde{y}$ that assigns the original ground-truth class to every pixel within the preserve mask $M$ and the dataset ignore index to every generated pixel,
\begin{equation}
    \tilde{y}_{i,j} \;=\;
    \begin{cases}
        y_{i,j} & \text{if } M_{i,j} = 1, \\
        \iota & \text{otherwise}.
    \end{cases}
    \label{eq:syn-label}
\end{equation}
Together, \cref{eq:syn-label,eq:paste-back,eq:loss} ensure label validity throughout training: \cref{eq:paste-back} restores the exact RGB values of every supervised pixel, \cref{eq:syn-label} assigns valid labels only to those restored pixels and marks all generated pixels as ignore, and \cref{eq:loss} excludes ignored pixels from the loss. Consequently, synthetic context influences the training through patch-wise self-attention, but generated pixels never serve as direct training targets.

\subsection{Iterative Segmenter Refinement}
\label{sec:method:train}

Finally, we propose to improve the baseline segmenter $f_{\theta_0}$ by fine-tuning it on $\mathcal{D} \cup \mathcal{D}_{\mathrm{syn}}$, which now specifically contains samples that are challenging for $f_{\theta_0}$.
One iteration of our approach is shown in Algorithm \ref{alg:method}. Our approach can be repeated with the improved segmenter $f_{\theta_1}$: recompute entropy, select the currently most informative regions, synthesize a new $\mathcal{D}_{\mathrm{syn}}$, and fine-tune again. 
This yields an active-learning-style sequence $f_{\theta_0}, f_{\theta_1}, \ldots, f_{\theta_T}$ in which each round automatically targets the latest model's remaining failure modes, until performance saturates.

\section{Experiments}
\label{sec:experiments}

\subsection{Experimental Setup}
\label{sec:exp:setup}

\vspace{6px}

\noindent \textbf{Datasets.} We evaluate on Cityscapes~\cite{cordts2016cityscapes} (urban driving, 19 semantic classes, 2975 training / 500 validation images at $1024 \times 2048$), UAVID~\cite{lyu2020uavid} (aerial urban scenes, 8 classes, $\sim$200 training / 70 validation images at $4096 \times 2160$), and BDD100K~\cite{yu2020bdd100k} (urban driving in diverse weather and times of day, 19 classes, 7000 training / 1000 validation images at $720 \times 1280$). The three datasets span different viewpoints (street-level vs.\ aerial), different geographic distributions, and different scene complexities, providing a more stringent test of generalization than any single benchmark. 
Similar to our baselines \cite{li2024simple,kupyn2024dataset}, we utilize fractions of the training data to evaluate low-data scenarios, but always use all available validation samples.

\begin{table*}[t]
    \centering
    \begin{minipage}[t]{0.68\linewidth}
        \caption{
        \textbf{Main results on Cityscapes 10\%} (mean $\pm$ std over five seeds), with representative rare and frequent classes by training-pixel share. Improvements concentrate on rare classes.
        }
        \label{tab:main_cityscapes}
    \end{minipage}%
    \hfill
    \begin{minipage}[t]{0.28\linewidth}
        \caption{\textbf{Main results on BDD100K 10\%} (mean $\pm$ std over five seeds).}
        \label{tab:main_bdd}
    \end{minipage}
    \vspace{0.2cm}
    \begin{minipage}[t]{0.68\linewidth}
        \centering
        \setlength{\tabcolsep}{4pt}
        \small
        \resizebox{\linewidth}{!}{%
        \begin{tabular}[t]{l c c c c c c c}
        \toprule
        & & \multicolumn{3}{c}{Rare Classes} & \multicolumn{3}{c}{Frequent Classes} \\
        \cmidrule(lr){3-5}\cmidrule(lr){6-8}
        Method & mIoU & truck & bus & train & road & sidewalk & vegetation \\
        \midrule
        Real data only                         & $69.60$ & $72.80$ & $56.83$ & $56.04$ & $\underline{97.93}$ & $82.25$ & $90.86$ \\
        Simple Background Augmentation~\cite{li2024simple}            & $70.29 \pm 0.43$ & $70.36$ & $58.64$ & $56.63$ & $97.85$ & $81.71$ & $\mathbf{90.96}$ \\
        Instance Augmentation~\cite{kupyn2024dataset}      & $70.71 \pm 0.55$ & $73.66$ & $57.16$ & $56.03$ & $\mathbf{98.00}$ & $\mathbf{82.40}$ & $\underline{90.96}$ \\
        \midrule
        Ours, iteration 1                          & $71.67 \pm 0.45$ & $\mathbf{77.80}$ & $\underline{68.04}$ & $57.12$ & $97.90$ & $82.03$ & $90.59$ \\
        Ours, iteration 2                          & $\underline{72.09 \pm 0.67}$ & $77.02$ & $67.82$ & $\underline{62.59}$ & $97.91$ & $82.08$ & $90.68$ \\
        Ours, iteration 3                          & $\mathbf{72.24 \pm 0.18}$ & $\underline{77.47}$ & $\mathbf{68.77}$ & $\mathbf{65.28}$ & $97.92$ & $\underline{82.26}$ & $90.78$ \\
        \bottomrule
        \end{tabular}%
        }
    \end{minipage}%
    \hfill
    \begin{minipage}[t]{0.28\linewidth}
        \centering
        \renewcommand{\arraystretch}{0.91}
        \setlength{\tabcolsep}{6pt}
        \small
        \begin{tabular}[t]{l c}
        \toprule
        Method & mIoU \\
        \midrule
        Real data only                         & $59.60$ \\
        Simple B. A.~\cite{li2024simple}            & $59.72 \pm 0.25$ \\
        Instance Aug.~\cite{kupyn2024dataset}      & $60.45 \pm 0.13$ \\
        \midrule
        Ours, iteration 1                          & $60.57 \pm 0.14$ \\
        Ours, iteration 2                          & $\underline{61.00 \pm 0.26}$ \\
        Ours, iteration 3                          & $\mathbf{61.17 \pm 0.16}$ \\
        \bottomrule
        \end{tabular}
    \end{minipage}
\end{table*}

\begin{table*}[t]
\centering
\caption{
\textbf{Main results on UAVID 100\%} (mean $\pm$ std over five seeds), with all individual class results. Our method achieves the best results for every class and is particularly effective for rare classes by training-pixel share.
}
\label{tab:main_uavid}
\setlength{\tabcolsep}{4pt}
\small
\resizebox{0.9\textwidth}{!}{%
\begin{tabular}{l c c c c c c c c}
\toprule
& & \multicolumn{3}{c}{Rare Classes} & \multicolumn{4}{c}{Frequent Classes} \\
\cmidrule(lr){3-5}\cmidrule(lr){6-9}
Method & mIoU & moving\_car & static\_car & human & building & road & tree & low\_veg \\
\midrule
Real data only                         & $60.31$ & $57.05$ & $59.58$ & $28.72$ & $94.98$ & $87.32$ & $81.18$ & $73.66$ \\
Simple Background Augmentation~\cite{li2024simple}            & $61.49 \pm 0.35$ & $58.99$ & $63.81$ & $31.61$ & $95.06$ & $87.19$ & $81.01$ & $74.25$ \\
Instance Augmentation~\cite{kupyn2024dataset}      & $62.24 \pm 0.25$ & $60.79$ & $64.84$ & $\underline{33.24}$ & $95.52$ & $88.09$ & $81.35$ & $74.10$ \\
\midrule
Ours, iteration 1                          & $62.58 \pm 0.45$ & $62.47$ & $64.90$ & $31.38$ & $95.42$ & $88.09$ & $82.47$ & $75.94$ \\
Ours, iteration 2                          & $\underline{63.48 \pm 0.34}$ & $\underline{64.57}$ & $\underline{68.19}$ & $32.72$ & $\mathbf{95.56}$ & $\underline{88.47}$ & $\mathbf{83.06}$ & $\mathbf{77.20}$ \\
Ours, iteration 3                          & $\mathbf{63.99 \pm 0.22}$ & $\mathbf{65.80}$ & $\mathbf{69.28}$ & $\mathbf{33.32}$ & $\underline{95.54}$ & $\mathbf{88.61}$ & $\underline{82.89}$ & $\underline{77.08}$ \\
\bottomrule
\end{tabular}
}
\end{table*}

\vspace{6px}

\noindent \textbf{Implementation Details.} For segmentation, we use a DINOv2~\cite{oquab2023dinov2} pre-trained ViT~\cite{dosovitskiy2020image} backbone paired with a linear semantic decoder. Both the real-data baseline checkpoint $\theta_{0}$ and the fine-tuned segmenters $\theta_{1-3}$ are trained with AdamW~\cite{loshchilov2017decoupled} and a cosine learning-rate schedule. For all experiments, we report mean validation mIoU with standard deviation over 5 seeds. Fine-tuning starts from $\theta_{0}$ on the union $\mathcal{D} \cup \mathcal{D}_{\mathrm{syn}}$. For inpainting, we use a black-box latent diffusion inpainter with 40 denoising steps, classifier-free guidance scale 7.0, and no ControlNet or auxiliary conditioning. We choose SDXL-Inpaint-1.0~\cite{podell2024sdxl} for a fair comparison with baselines~\cite{kupyn2024dataset, li2024simple}, but note that any off-the-shelf inpainting model can be used. Cityscapes images are randomly cropped to $1024 \times 1024$; UAVID images are processed by cropping to a random $2048\times2048$ region, bilinear-resizing to $1024 \times 1024$ for generation, and upsampling back; BDD100K is processed at $1024 \times 1024$ after letterbox padding. The text prompt is a single dataset-level descriptor (e.g.\ \texttt{photorealistic, ultra-detailed,[...] in the style of the Cityscapes dataset}, see Appendix \ref{sec:supp_prompts}).
We run our approach for a maximum of three iterations since we empirically observed saturation after more iterations.
If not stated otherwise, we use preserve region fraction $\tau = 0.10$.

\vspace{6px}

\noindent \textbf{Baselines.} We compare against:
(i) the no-synthetic-data baseline trained only on real data $\mathcal{D}$;
(ii) \emph{Instance Augmentation}~\cite{kupyn2024dataset}, which regenerates all foreground object instances using a stack of auxiliary models;
(iii) \emph{Simple Background Augmentation}~\cite{li2024simple}, which regenerates the entire background given a set of foreground objects.
Here, we assign foreground/background based on classes (e.g. \textit{car} as foreground and \textit{wall} as background, see Appendix \ref{app:fg_bg_split}).

\vspace{6px}

\noindent \textbf{Code.} \url{https://github.com/XITASO/Preserve-the-Hard-Regenerate-the-Rest} 

\subsection{Main Results}
\label{sec:exp:main}

\Cref{tab:main_cityscapes,tab:main_uavid,tab:main_bdd} report our main results against both spatial preserve-and-regenerate augmentation baselines on the three datasets.
On all datasets, we find consistent gains over the no-augmentation reference ($+3.68 / +2.64 / +1.57$ mIoU on UAVID / Cityscapes / BDD) vs. only $+1.93 / +1.11 / +0.85$ for the best baseline. Notably, we more than double the gain over the strongest baseline on Cityscapes. 
A per-class breakdown verifies the intuition that classes that are rare by proportion of annotated training pixels benefit most from our method: we obtain up to $+11.94 / +9.24 / +5.00$ IoU on bus / train / truck (Cityscapes, \cref{tab:main_cityscapes}) and up to $+8.75 / +9.70 / +4.60$ IoU on moving\_car / static\_car / human (UAVID, \cref{tab:main_uavid}). 

\subsection{Effect of Training-Set Size}
\label{sec:exp:datasize}

Data scarcity is central to our motivation, so we study how the benefit of uncertainty-guided augmentation scales with the amount of available real data. \Cref{tab:datasweep} reports Cityscapes results for train splits of 5\%, 10\%, 33\%, and 100\%, comparing the real-only baseline, the strongest spatial baseline (Instance Augmentation~\cite{kupyn2024dataset}), and our method. The gain is consistent across all settings and remains substantially ahead of the baseline across the full range.

\subsection{Ablation Studies}
\label{sec:exp:ablation}

\vspace{6px}

\noindent \textbf{Region Selection and Inpainting.} The central claim of our method is that the \emph{informational} axis, not the spatial one, is the right axis along which to allocate synthetic context. To test this, we keep the entire downstream pipeline (inpainting, paste-back, ignore-mask, fine-tuning) identical, varying only the rule that selects \emph{which} region is preserved (\cref{tab:ablation_selection}): (i) a random square crop of area $\tau$; (ii) randomly chosen ground-truth classes summing to area $\tau$; (iii) the most-\emph{confident} classes, i.e.\ the inverse of our criterion; (iv) our most-uncertain classes but \emph{without} inpainting --- the original context is kept and the loss is simply restricted to the preserved region; and (v) our full method, most-uncertain classes with inpainted context. We find that selecting most-confident classes yields near-zero gain, and that region selection by random cropping or random class selection significantly falls behind our method. Also, we find that inpainting accounts for a large portion of the achieved gains and is necessary to place hard regions in novel visual context.

\begin{table}[t]
\centering
\caption{\textbf{Effect of training-set size} on Cityscapes (5 seed mean validation mIoU $\pm$ std).}
\label{tab:datasweep}
\setlength{\tabcolsep}{5pt}
\small
\resizebox{\columnwidth}{!}{%
\begin{tabular}{l c c c c c}
\toprule
Split & Real only & Instance Aug.~\cite{kupyn2024dataset} & Ours, iteration 1 & Ours, iteration 2 & Ours, iteration 3 \\
\midrule
5\%   & $61.42$    & $62.37 \pm 0.14$             & $64.14 \pm 0.18$                      & $64.23  \pm 0.11$       & $\mathbf{64.64 \pm 0.18}$ \\
10\%  & $69.60$    & $70.71 \pm 0.55$             & $71.67 \pm 0.45$                      & $72.09  \pm 0.67$       & $\mathbf{72.24  \pm 0.18}$ \\
33\%  & $72.54$    &      $75.05 \pm 0.18$            & $75.44  \pm 0.21$                     & $76.36  \pm 0.06$       & $\mathbf{76.77  \pm 0.16}$ \\
100\% & $75.46$    & $76.78 \pm 0.14$             & $76.85 \pm 0.21$                      & $77.38  \pm 0.24$       & $\mathbf{77.69  \pm 0.22}$ \\
\bottomrule
\end{tabular}%
}
\end{table}

\begin{table}[t]
\centering
\caption{\textbf{Region-selection ablation} on Cityscapes 10\% (5 seed validation mIoU $\pm$ std, single augmentation iteration).}
\label{tab:ablation_selection}
\setlength{\tabcolsep}{4pt}
\small
\resizebox{\columnwidth}{!}{%
\begin{tabular}{l c c c c}
\toprule
Selection rule & mIoU & truck & bus & train \\
\midrule
Real data only                              & $69.60$ & $72.80$ & $56.83$ & $56.04$ \\
Confident classes ($\tau$)                     & $70.06 \pm 0.35$ & $74.19$ & $53.74$ & $59.45$ \\
Random square region ($\tau$)                  & $70.53 \pm 0.42$ & $74.80$ & $54.29$ & $\mathbf{62.81}$ \\
Random classes ($\tau$)                        & $\underline{70.61 \pm 0.20}$ & $\underline{75.08}$ & $59.69$ & $\underline{59.89}$ \\
\midrule
Uncertain classes, no inpainting               & $70.13 \pm 0.11$ & $71.45$ & $\underline{63.66}$ & $56.52$ \\

\midrule
\textbf{Uncertain classes + inpainting (Ours)} & $\mathbf{71.67 \pm 0.45}$ & $\mathbf{77.80}$ & $\mathbf{68.04}$ & $57.12$ \\
\bottomrule
\end{tabular}%
}
\end{table}
\vspace{6px}

\noindent \textbf{Paste-Back \& Ignore-Mask.} In combination, the paste-back step (\cref{eq:paste-back}) and the ignore-mask (\cref{eq:syn-label}) guarantee that no label noise is introduced. To isolate their effect, we compare all four possible configurations: with and without original-pixel paste-back, and with and without the synthetic-pixel ignore mask applied to the inpainted region. \Cref{tab:ablation_pasteback} shows that paste-back and ignore-mask contribute roughly equally, but that both fall behind our full method.
\vspace{6px}

\noindent \textbf{Preserve-Area Threshold $\mathbf{\tau}$.} The threshold $\tau$ controls the trade-off between supervision density (more preserved pixels per image) and contextual diversity (more pixels regenerated). \Cref{tab:ablation_tau} sweeps $\tau \in \{0.05, 0.10, 0.15, 0.20, 0.25\}$. Gains occur across the full range, with the best setting at $\tau = 0.10$. This suggests that selecting a small set of high-uncertainty regions yields a greater benefit than coarsely selecting uncertain regions.

\begin{table}[b]
\centering
\begin{minipage}[t]{0.46\columnwidth}
\centering
\caption{\textbf{Effect of eliminating label noise} (Cityscapes 10\%). Both ignore mask and pixel-exact paste-back contribute.}
\label{tab:ablation_pasteback}
\setlength{\tabcolsep}{3pt}
\resizebox{0.95\linewidth}{!}{%
\begin{tabular}{ccc}
\toprule
Ignore & Paste-back & mIoU \\
\midrule
\xmark & \xmark & $69.10 \pm 0.62$ \\
\xmark & \cmark & $70.74 \pm 0.33$ \\
\cmark & \xmark & $70.98 \pm 0.29$ \\
\cmark & \cmark & $\mathbf{71.67 \pm 0.45}$ \\
\bottomrule
\end{tabular}}
\end{minipage}
\hfill
\begin{minipage}[t]{0.50\columnwidth}
\centering
\caption{\textbf{Generalization across backbone and inpainter} (Cityscapes 10\%, mIoU $\pm$ 5 seed std, single iteration).}
\label{tab:generalization}
\setlength{\tabcolsep}{3pt}
\resizebox{\linewidth}{!}{%
\begin{tabular}{llcc}
\toprule
Backbone & Inpaint & Base & Ours  \\
\midrule
ViT (DINOv2)       & SDXL & 69.6 &  $71.7 \pm 0.5
$ \\
ViT (DINOv2)        & FLUX & 69.6 & $\mathbf{71.9 \pm 0.4}$  \\

\midrule
SegFormer & SDXL & 67.3 & $\mathbf{68.6 \pm 0.1}$  \\ 
\bottomrule
\end{tabular}}
\end{minipage}
\end{table}

\vspace{6px}

\noindent \textbf{Real-to-Synthetic Sampling Ratio.} \Cref{tab:ablation_ratio} sweeps the fine-tuning ratio of synthetic to real samples. 1:1 is best, i.e. our method is most useful as a substantial complement.

\begin{table}[t]
\centering
\caption{\textbf{Preserve-area threshold sweep} on Cityscapes 10\% (5 seed mean validation mIoU $\pm$ standard deviation, single iteration).}
\label{tab:ablation_tau}
\resizebox{\columnwidth}{!}{%
\begin{tabular}{l c c c c c}
\toprule
$\tau$ & $0.05$ & $\mathbf{0.10}$ & $0.15$ & $0.20$ & $0.25$ \\
\midrule
mIoU & $71.00 \pm 0.40$ & $\mathbf{71.67 \pm 0.45}$ & $\underline{71.29 \pm 0.15}$ & $71.16 \pm 0.28$ & $70.54 \pm 0.32$ \\
\bottomrule
\end{tabular}%
}
\end{table}

\begin{table}[t]
\centering
\caption{\textbf{Synthetic-to-real sampling ratio} on Cityscapes 10\% (5 seed mean validation mIoU $\pm$ standard deviation, single iteration).}
\label{tab:ablation_ratio}
\setlength{\tabcolsep}{5pt}
\small
\resizebox{\columnwidth}{!}{%
\begin{tabular}{l c c c c c}
\toprule
Syn\,:\,Real & 0:1 (baseline) & 1:2 & $\mathbf{1:1}$ & 1.5:1 & 2:1 \\
\midrule
mIoU & $69.60$ & $71.40 \pm 0.59$ & $\mathbf{71.67 \pm 0.45}$ & $\underline{71.44 \pm 0.14}$ & $71.28 \pm 0.31$ \\
\bottomrule
\end{tabular}
}
\end{table}

\subsubsection{Compute Analysis}
\label{sec:exp:abl:compute}

\Cref{tab:compute} reports per-sample wall-clock time on a single NVIDIA A100 40GB for the full pipeline. The cost of our method is dominated by the SDXL inpainting forward pass; the uncertainty-based region selection, the paste-back, and the label construction are essentially free. This means that the additional uncertainty-based component we add over a vanilla inpainting pipeline carries negligible overhead.

\begin{table}[b]
\centering
\caption{\textbf{Per-sample wall-clock breakdown} of the augmentation pipeline on a single NVIDIA A100 40GB at $1024 \times 1024$ generation resolution. Region selection contributes negligible overhead relative to the SDXL inpainting forward pass.}
\label{tab:compute}
\setlength{\tabcolsep}{6pt}
\small
\begin{tabular}{l r r}
\toprule
Operation & Time (s) & \% of total \\
\midrule
Segmenter forward (entropy)      & 0.026 & 0.42 \\
Per-class entropy aggregation    & 0.003 & 0.04 \\
Greedy class selection           & 0.000 & $<$0.01 \\
\midrule
\emph{Region selection subtotal} & 0.029 & 0.46 \\
\midrule
SDXL inpainting (40 steps)       & 6.230 & 98.77 \\
Paste-back \& label construction & 0.049 & 0.77 \\
\midrule
\textbf{Total per sample}        & 6.308 & 100.00 \\
\bottomrule
\end{tabular}
\end{table}

\begin{figure*}[t]
\centering
\begin{subfigure}{\linewidth}
    \centering
    \includegraphics[width=\linewidth]{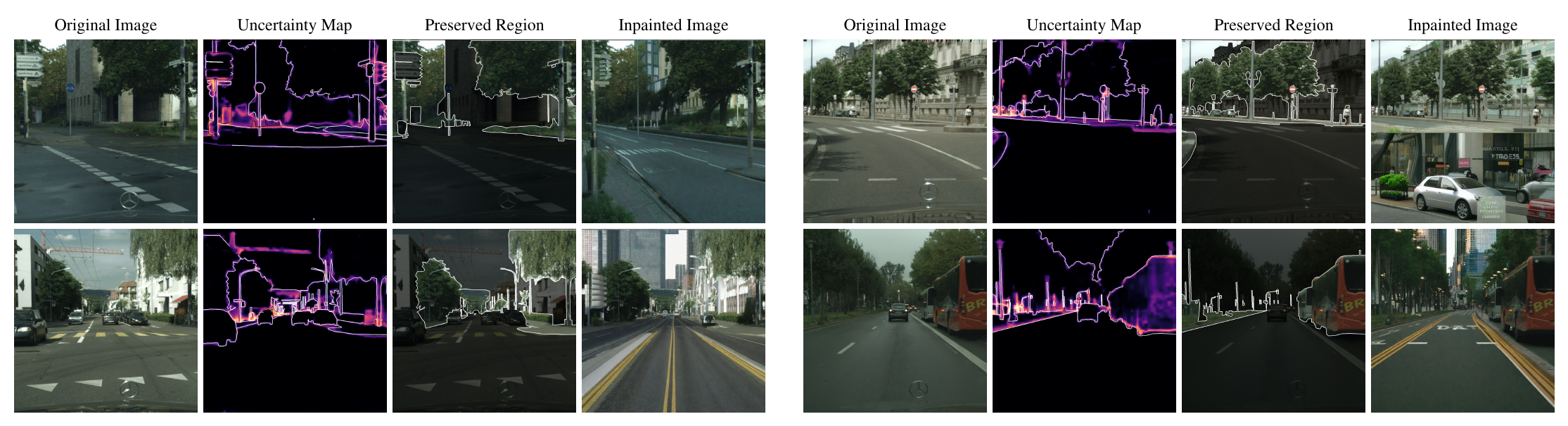}
    \caption{Cityscapes (street-level driving)}
    \label{fig:qualitative_cityscapes}
\end{subfigure}

\vspace{4pt}

\begin{subfigure}{\linewidth}
    \centering
    \includegraphics[width=\linewidth]{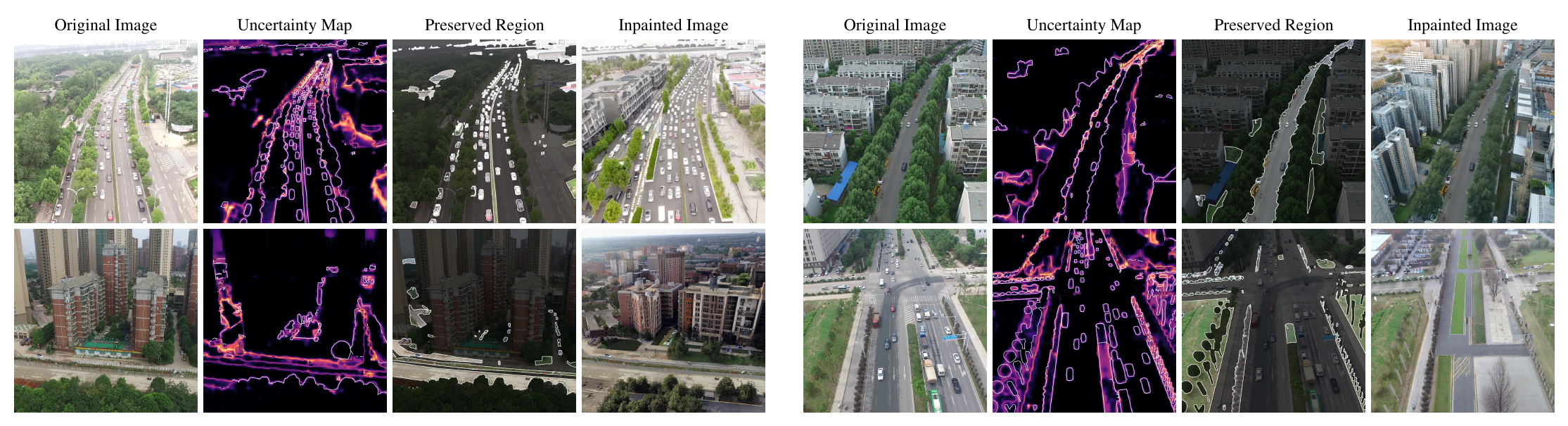}
    \caption{UAVID (aerial urban scenes)}
    \label{fig:qualitative_uavid}
\end{subfigure}

\caption{\textbf{Uncurated qualitative results across both domains.} For each example, the columns show the original image, the baseline segmenter's per-pixel predictive-entropy map, the preserve mask formed from the most uncertain classes, and the final augmented sample after inpainting and pixel-exact paste-back. Generation uses dataset-level prompts: \textit{``photorealistic, ultra-detailed, 4K high-resolution, sharp focus, high quality, in the style of the Cityscapes dataset''} for Cityscapes and \textit{``photorealistic aerial drone image, urban scene, high-resolution, realistic lighting, sharp overhead perspective''} for UAVID. In both domains the preserved pixels remain bitwise identical to the source, so context diversity is injected without perturbing a single supervised pixel; samples are drawn at random without curation and a larger quantity of uncurated samples is provided in the Appendix.}

\label{fig:qualitative}
\end{figure*}


\subsection{Generalization Across Backbone and Inpainter}
\label{sec:exp:general}

To confirm that our gains are not tied to a single architecture, we vary the backbone and the inpainter independently (Tab.~\ref{tab:generalization}). Swapping the DINOv2 ViT for a SegFormer \cite{xie2021segformer} backbone preserves the improvement, indicating that the method is segmenter-agnostic. Replacing the SDXL inpainter with FLUX \cite{blackforestlabs2024flux1filldev} yields marginal gain, suggesting that any sufficiently capable off-the-shelf inpainter suffices.

\subsection{Qualitative Results}
\label{sec:exp:qual}

\Cref{fig:qualitative_cityscapes,fig:qualitative_uavid} show uncurated synthetic samples and we provide more uncurated samples for Cityscapes and UAVID in Appendix \ref{sec:supp_qual}. The preserved region remains identical, while the surrounding scene provides novel context, e.g., different lighting, season, and street layout. Although artifacts can occur in the inpainted regions, those pixels are excluded from the loss by definition.

\vspace{6px}

\noindent \textbf{Limitations.} Our method requires (i) the baseline segmenter to compute entropy at every iteration and (ii) an SDXL-class diffusion inpainter for synthetic sample generation. The compute cost is dominated by the inpainter (\cref{tab:compute}).
On domains where SDXL has weak priors (e.g.\ medical imaging) we expect consistent but smaller gains. 
Domain-specific inpainters could mitigate this problem.

\section{Conclusion}
\label{sec:conclusion}

This paper revisited preserve-and-regenerate augmentation for semantic segmentation and showed that allocating the synthetic-data budget by model uncertainty is more effective than using fixed spatial rules such as foreground or background regeneration.
Across three street-level and aerial benchmarks, our method substantially improves segmenter performance over real-data training and outperforms spatial preserve-and-regenerate baselines.
The improvements are largest for challenging classes with underrepresented or small objects with few annotated pixels.
Our ablations demonstrate that all components of the approach are necessary and contribute to the results: 
Uncertainty-based region selection, context inpainting and pixel-exact paste-back combined with an ignore-mask loss each contribute to the final gain. 
These results suggest a clear use-case: our proposed method is most beneficial when segmentation performance is limited by scarce or imbalanced dense annotations, and especially when the challenging classes occupy few pixels. 
It is particularly suitable for saftey-critical settings where synthetic augmentation must not corrupt the labels and labeled synthetic data is difficult to trust.

Our method is deliberately simple, resource effective and architecture-agnostic, as we verify by transferring it across segmenter backbones and inpainters.
Rather than relying on fully synthetic data or deciding with which spatial heuristic to inpaint, one can simply \emph{preserve what is hard and regenerate the rest}.

\paragraph{Acknowledgements}
This work was carried out in the context of the ENGEL project, funded from the Federal Ministry of Economic Affairs and Energy (BMWE) of Germany through the Federal Aviation Research Program (LuFo) VI-3, under FKZ 20F2201D.

This paper is also supported by the DAAD programme Konrad Zuse Schools of Excellence in Artificial Intelligence, sponsored by the Federal Ministry of Research, Technology and Space.

{
    \small
    \bibliographystyle{ieeenat_fullname}
    \bibliography{main}
}

\clearpage

\twocolumn[{%
  \begin{center}
      \Large \textbf{Appendix}
      \vspace{0.75cm} 
  \end{center}
}]

\begin{figure*}[ht]
\centering
\includegraphics[width=\textwidth]{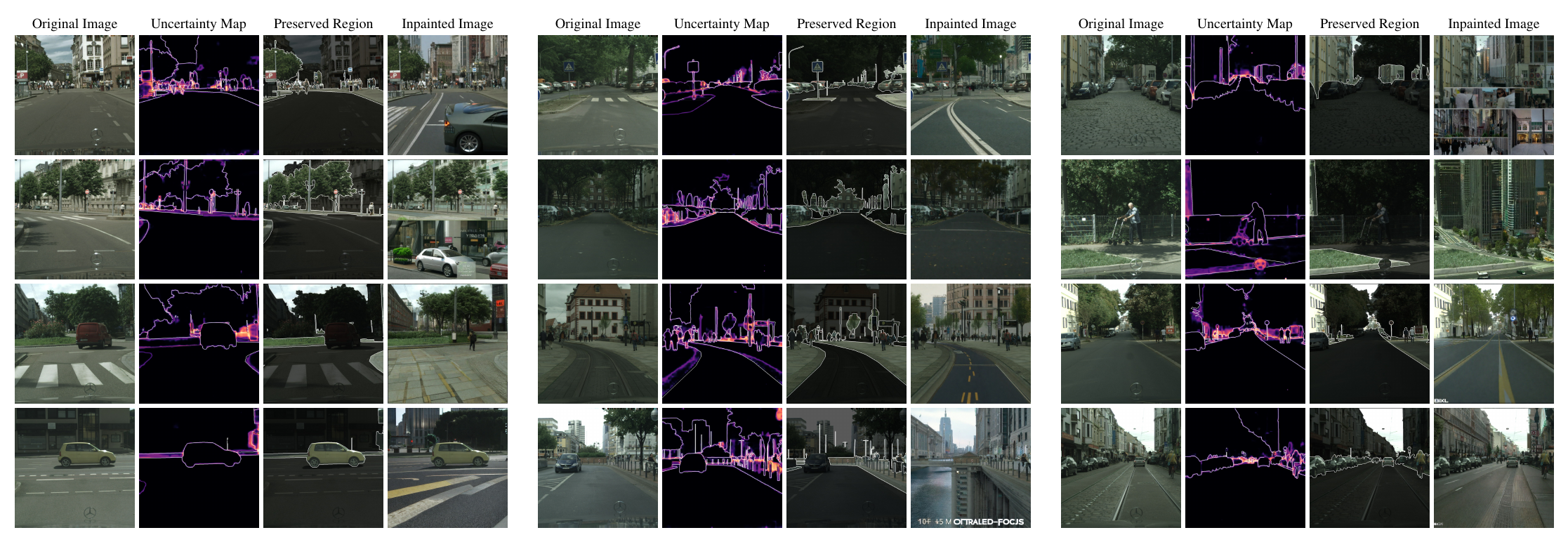}
\caption{Additional uncurated qualitative samples on \textbf{Cityscapes}
(street-level driving).}
\label{fig:supp_cs}
\end{figure*}

\begin{figure*}[ht]
\centering
\includegraphics[width=\textwidth]{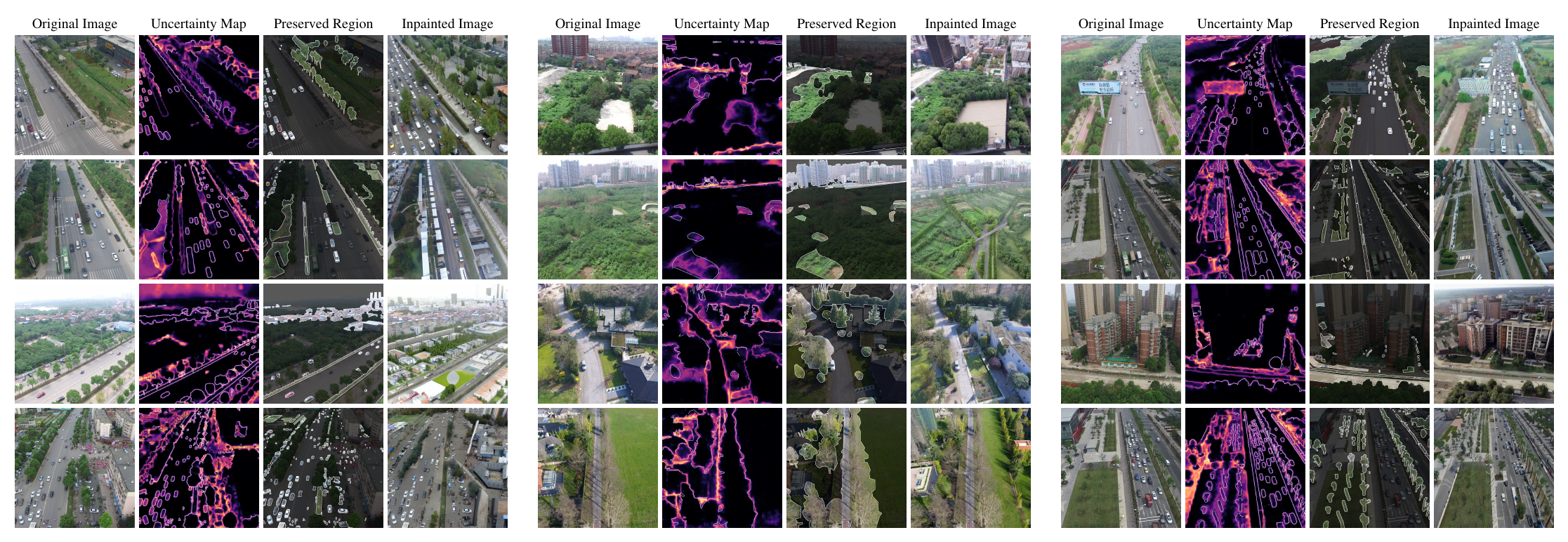}
\caption{Additional uncurated qualitative samples on \textbf{UAVID}
(aerial urban scenes).}
\label{fig:supp_uavid}
\end{figure*}

\section{Additional Uncurated Qualitative Samples}
\label{sec:supp_qual}

We show more examples that are sampled without any qualitative filtering in \Cref{fig:supp_cs} and \Cref{fig:supp_uavid}. The columns
show the original image, the predictive-entropy map, the preserved uncertain
region, and the final augmented image after inpainting and paste-back. Every
preserved pixel in the augmented image is bit-wise identical to the source.

\section{Additional Implementation Details}
\label{sec:supp_impl}

This section specifies everything needed to reproduce the method and results
in the main paper: the fixed data subsets, preprocessing, the training and
fine-tuning schedules, generation settings, and the checkpoint-selection
rule.

\begin{table}[b]
\centering
\small
\begin{tabular}{@{}p{0.42\columnwidth}p{0.52\columnwidth}@{}}
\toprule
\textbf{Setting} & \textbf{Value} \\
\midrule
Backbone & DINOv2 ViT \\
Decoder & Linear semantic decoder \\
Encoder update & Trainable; LR multiplier 0.1 \\
Training crop & $1024\times1024$ \\
Optimizer & AdamW \\
Decoder peak LR & $1\times10^{-4}$ \\
Encoder peak LR & $1\times10^{-5}$ \\
Loss & Ignore-index masked Cross-entropy \\
LR schedule & Cosine decay \\
Warmup & None \\
Weight decay & 0.05 \\
Batch size & 4 per GPU \\
Gradient accumulation & 4 steps (effective batch 16) \\
Precision & 16-bit mixed precision \\
Maximum length & 20,000 optimizer steps \\
Validation interval & Every 1,000 optimizer steps \\
Data-loader workers & 16, pinned memory \\
Hardware & A100 40GB \\
\bottomrule
\end{tabular}
\caption{Real-only baseline training configuration.}
\label{tab:supp_seg}
\end{table}

\begin{table}[b]
\centering
\small
\begin{tabular}{@{}p{0.42\columnwidth}p{0.52\columnwidth}@{}}
\toprule
\textbf{Setting} & \textbf{Value} \\
\midrule
Initialization & $\theta_{0}$ \\
Training set & $\mathcal{D}\cup\mathcal{D}_{\mathrm{syn}}$ \\
Synthetic : real ratio & 1:1 \\
Decoder peak LR & $2\times10^{-5}$ \\
Encoder peak LR & $2\times10^{-6}$ \\
LR schedule & Cosine decay; restarted \\
Maximum length & 20,000 optimizer steps \\
Validation interval & Every 1,000 optimizer steps \\
Encoder update & Trainable \\
Other settings & Same as real-only training \\
\bottomrule
\end{tabular}
\caption{Fine-tuning configuration for the augmented training iterations. The
optimizer and LR scheduler are reinitialized each iteration.}
\label{tab:supp_ft}
\end{table}

\subsection{Data Splits and Reproducibility}
\label{sec:supp_data}

The training data splits used in our experiments are fixed independently of the training seed. 
For each dataset $\mathcal{D}$, we sort all pairs of image and label map $(x,y)\in \mathcal{D}$ by filename, shuffle once, and keep the first $\lfloor\rho N\rfloor$ pairs for split fraction
$\rho$ (e.g., $\rho=0.1$ for the 10\% Cityscapes split).
The different training seeds that we used for mean and standard deviation results therefore control model initialization, data-loader order, augmentation randomness, optimizer behaviour etc..
They do not change \emph{which} images are used: baseline training, preserve region selection, synthetic generation, and fine-tuning all draw from the same fixed training image IDs.

\paragraph{Dataset-specific preprocessing.}
Generation for all datasets operates on $1024\times1024$ pixel images. Cityscapes crops are taken from the
native $1024\times2048$ frame; BDD100K is zero-padded to $1024\times1024$. For UAVID, each native $4096\times2160$ frame is randomly cropped to a $2048\times2048$ image and then resized to $1024\times1024$ (bilinear for
RGB, nearest-neighbor for labels). 

\subsection{Segmentation Model and Baseline Training}
\label{sec:supp_seg}

The segmenter $f_{\theta}$ pairs the
DINOv2-pretrained ViT encoder with a linear per-patch decoder. The decoder is a single fully connected linear layer, mapping each DINOv2 ViT patch token from 384 channels directly to \(C\) class logits.
The encoder is trainable and both modules are optimized end to end
under the ignore-index cross-entropy loss. We report training details in \Cref{tab:supp_seg}.

\paragraph{Training augmentation.}
In addition to our proposed synthetic data augmentation, we also utilize simple train-time augmentation.
Each sample undergoes photometric jitter, a random horizontal flip and scale
jitter. 

\subsection{Fine-Tuning on \texorpdfstring{$\mathcal{D}\cup\mathcal{D}_{\mathrm{syn}}$}{D U D\_syn}}
\label{sec:supp_ft}

Fine-tuning starts from the real-data-only checkpoint $\theta_0$.
We concatenate the real dataset $\mathcal{D}$ and the synthetic dataset $\mathcal{D}_{\mathrm{syn}}$, then
shuffle them together. We report fine-tuning details in \Cref{tab:supp_ft}.

\subsection{Diffusion Inpainting and Generation}
\label{sec:supp_gen}

We use the public SDXL-Inpaint-1.0 checkpoint
(\texttt{diffusers/stable-diffusion-xl-1.0-inpai nting-0.1}) with no
ControlNet, classifier guidance, or auxiliary spatial conditioning. All cropped image inputs are square, so the SDXL aspect bucket is always
$1024\times1024$. In general the implementation selects the nearest
canonical SDXL bucket, resizes RGB bilinearly and labels/masks
by nearest-neighbor, and restores the cached resolution after generation. Each
dataset uses a single positive and single negative prompt, with no per-image
or per-class text.

\subsection{Generation Prompts}
\label{sec:supp_prompts}

\paragraph{Positive prompts.}
\begin{itemize}
\item \textbf{Cityscapes:}
``photorealistic, ultra-detailed, 4K high-resolution, sharp focus, high
quality, in the style of the Cityscapes dataset.''
\item \textbf{BDD100K:}
``photorealistic dashcam image, diverse urban driving scene,
high-resolution, realistic lighting and weather, sharp focus.''
\item \textbf{UAVID:}
``photorealistic aerial drone image, urban scene, high-resolution,
realistic lighting, sharp overhead perspective.''
\end{itemize}

\paragraph{Negative prompts.}
\begin{itemize}
\item \textbf{Cityscapes:}
``blurry, low quality, deformed, melted structures, floating objects,
cartoon, illustration, unrealistic shadows, out of perspective, wrong
scale.''
\item \textbf{BDD100K:}
``blurry, low quality, deformed vehicles, melted structures, cartoon,
illustration, unrealistic shadows, wrong scale, watermark, text.''
\item \textbf{UAVID:}
``blurry, low quality, cartoon, illustration, distorted vehicles, warped
roads, duplicated objects, unrealistic shadows, text, watermark.''
\end{itemize}

\subsection{Region Selection and Mask Construction}
\label{sec:supp_mask}

As described in the main paper, for each non-ignore ground-truth class present in the image we average the entropy over the class's pixels, sort classes by decreasing mean entropy, and add them until their union exceeds $\tau HW$ pixels.
The resulting preserve area can therefore exceed $\tau$ when the last added class is large. Further details:

\begin{itemize}
\item Equal-entropy ties break by ascending label ID (stable sort).
\item No minimum object-area filter is applied.
\item In contrast to baselines \cite{kupyn2024dataset, li2024simple}, no preserve-mask dilation or inpaint-mask erosion is applied.
\item Ignore-index pixels are excluded from the entropy calculation and cannot be
selected for the preserve mask.
\item For BDD100K, padding and cropping are done before entropy scoring. The resulting image size is $1024\times1024$ pixels.
\item UAVID entropy is computed on each $1024\times1024$ image; there is no tiling, overlap, blending, or full-frame aggregation.
\end{itemize}

\subsection{Validation and Reported Metrics}
\label{sec:supp_eval}

During evaluation, no random crop augmentation is applied. Each validation image
is resized isotropically such that both spatial dimensions are at least
$1024$ pixels; equivalently, the shorter side is brought to $1024$ if needed.
The resized image is then evaluated using overlapping $1024\times1024$ windows
along the longer dimension. Logits from overlapping windows are averaged, and
the resulting logit map is bilinearly resized back to the original image
resolution before computing IoU. Per-class IoU is computed from a single global
confusion matrix over the full validation set, excluding ignore pixels, and
mIoU is the unweighted mean over semantic classes.

We validate once before the first optimizer step, every 1,000 steps
thereafter, and at termination, retaining the highest-mIoU checkpoint. Across training seeds we report
the arithmetic mean and standard deviation of the per-seed best mIoU.

\subsection{Iterative Refinement Protocol}
\label{sec:supp_iter}

We run three fixed rounds of creating synthetic training data with our approach. Each round $t$ computes entropy with segmenter $f_{\theta_{t-1}}$, generates a fresh synthetic set from the fixed real image IDs, and then initializes fine-tuning from the round-$(t-1)$ checkpoint.
Earlier synthetic sets are discarded rather than accumulated.
The default fine-tuning LR is $2\times10^{-5}$ (decoder) and $2\times10^{-6}$ (encoder).


\section{Foreground/Background Split For Baseline Methods}
\label{app:fg_bg_split}

The two synthetic augmentation baselines used in our experiment rely on spatial heuristics based on foreground objects and the  background of each image.
To apply this to densely labeled semantic segmentation use-cases like Cityscapes \cite{cordts2016cityscapes}, we assign each semantic class to either foreground or background. The exact classes are listed in \Cref{tab:fg_bg_split}. 

\begin{table}[t]
\centering
\caption{Canonical foreground/background split used for baseline implementations.}
\label{tab:fg_bg_split}
\setlength{\tabcolsep}{4pt}
\small
\begin{tabular}{p{0.20\linewidth} p{0.33\linewidth} p{0.34\linewidth}}
\toprule
Dataset & Foreground & Background \\
\midrule
Cityscapes / BDD100K &
person, rider, car, truck, bus, train, motorcycle, bicycle &
road, sidewalk, building, wall, fence, pole, traffic light, traffic sign, vegetation, terrain, sky, ignore \\
\midrule
UAVID &
static car, human, moving car &
clutter, building, road, tree, vegetation, ignore \\
\bottomrule
\end{tabular}
\end{table}

\end{document}